\documentclass{article}

\usepackage{microtype}
\usepackage{graphicx}
\usepackage{subcaption}
\usepackage{booktabs} 
\usepackage[table]{xcolor}
\usepackage{hyperref}
\newcommand{\Ci}{C_i}

\usepackage[preprint]{icml2026}

\usepackage{amsmath}
\usepackage{amssymb}
\usepackage{mathtools}
\usepackage{amsthm}

\usepackage[capitalize,noabbrev]{cleveref}

\theoremstyle{plain}

\theoremstyle{definition}

\theoremstyle{remark}

\usepackage[textsize=tiny]{todonotes}

\icmltitlerunning{DocHRL: Hierarchical RL for Cost-Optimised Document Classification}

\begin{document}

\twocolumn[
  \icmltitle{\textsc{DocHRL}: A Hierarchical Reinforcement Learning\\
Framework for Cost-Optimised Document Classification}




  \begin{icmlauthorlist}
    \icmlauthor{Mohammed Yousif}{rm}
    \icmlauthor{Prabhjot Singh}{rm}
    \icmlauthor{Arjun Pankajakshan}{rm}
    \icmlauthor{Madhu Reddiboina}{rm}
  \end{icmlauthorlist}

  \icmlaffiliation{rm}{RediMinds, Inc., Detroit, Michigan, USA \\
\textmd{\small Email: \{firstname.lastname\}@rediminds.com}}

  \icmlcorrespondingauthor{ }

  \vskip 0.3in
]



\printAffiliationsAndNotice{}  

\begin{abstract}
Real-world document classification pipelines typically apply the same
sequence of models to every incoming document, regardless of its
complexity or type.  This leads to inefficient use of compute and human
resources: simple documents are over-processed while difficult ones may
not receive enough scrutiny.  We introduce \emph{DocHRL}, a hierarchical
reinforcement learning framework that learns to \emph{adaptively} and
\emph{dynamically} select the most cost-effective classification policy
on a per-document basis.  \emph{DocHRL} formulates document classification as
a sequential decision problem with a two-level policy hierarchy: a
top-level policy selects among broad options (vision classifiers, LLMs,
OCR, and human-in-the-loop review), while option-specific sub-policies
choose the concrete model or tool to invoke.  The reward signal is the
negative total expected cost, which captures inference cost, cost of
misclassification, and cost of human labelling.  Trained with Proximal
Policy Optimisation on the RVL-CDIP benchmark, \emph{DocHRL} achieves a macro
F1 of 0.973 across 16 document classes while reducing average per-document
cost to 2.74 normalised units compared to substantially higher costs
incurred by fixed standalone classifiers.  Our results demonstrate that
cost-aware reinforcement learning can simultaneously improve classification
performance and operational efficiency in document understanding systems.
\end{abstract}

\section{Introduction}
\label{sec:intro}
 
Document classification is the task of assigning one or more labels from
a predefined set to a given document.  It sits at the core of enterprise
document processing pipelines, where documents arrive across diverse
modalities including scanned images, parsed text, and structural
metadata.  In general, document classification is an information rich problem because a document can be represented across various modalities like \emph{image} (scanned image of a document), \emph{text} (parsed text from a document), and other \emph{metadata} (like the layout of a document). Consequently, document classification has been extensively addressed using image-based methods~\cite{afzal2015deepdocclassifier, harley2015evaluation, ferrando2020improving, siddiqui2022deep}, parsing a document using Optical Character Recognition (OCR) plus classifier based methods~\cite{smith2007overview}, layout detection based methods~\cite{binmakhashen2019document}, Pretrained Language Models (PLMs) based methods~\cite{adhikari2019docbert}, multimodal embedding based methods~\cite{audebert2019multimodal, radford2021learning, xu2021layoutlmv2}, and finally Vision-Language Models (VLMs) based methods~\cite{chen2024internvl}. 

A practical gap shared by all of these approaches is that they treat each
document with the same computational effort, regardless of how easy or hard
it is to classify.  In production, a pipeline may chain several such
classifiers together with a human review step, leading to a total
operational cost that is \emph{fixed} and \emph{independent of the document
type}.  Over large volumes, even small per-document inefficiencies
accumulate into substantial costs; moreover, resource allocation is
suboptimal in both directions: simple documents consume more resources than
they need, while borderline cases may not receive adequate attention.
 
To address this, we propose \emph{DocHRL}, a \textbf{Doc}ument classification
framework based on \textbf{H}ierarchical \textbf{R}einforcement
\textbf{L}earning.  \emph{DocHRL} formulates the classification pipeline as a
Markov Decision Process whose reward is the \emph{negative total expected
cost} of a classification episode.  A two-level policy hierarchy enables
structured exploration over a heterogeneous set of classification
tools.  By including the cost of misclassification and human-in-the-loop
review directly in the reward function, \emph{DocHRL} jointly optimises for
accuracy and cost.

The main contributions of this paper are:                                                                                                                   
\begin{enumerate}                                                             
    \item A formal cost model for four document classification pipeline architectures, unifying inference cost, cost of failure, and cost of human labelling into a single objective.                                                                                                           
    \item A hierarchical RL framework for adaptive, cost-aware document classification, trained end-to-end with a novel PPO variant featuring a smooth tanh-gated trust region, bounded value head, and adaptive return normalization (AdaPolyTan).                                                                                                       
    \item Empirical evaluation on the RVL-CDIP benchmark demonstrates that \emph{DocHRL} reduces average per-document cost while achieving superior classification performance to all static baselines considered (standalone classifiers, ensembles, LLMs, and human-in-loop systems) across 16 document classes.
\end{enumerate}

\section{Related Work}
\label{sec:related}
 
\paragraph{Document classification.}
The RVL-CDIP benchmark~\cite{harley2015evaluation} has been the standard
evaluation testbed for document classification since its introduction.
Approaches range from CNN-based image classifiers to layout-aware
transformers such as LayoutLMv2~\cite{xu2021layoutlmv2} and document
understanding models such as Donut~\cite{kim2022donut}.  While these
methods continue to push accuracy, they uniformly ignore operational cost.
 
\paragraph{Cost-sensitive learning.}
Cost-sensitive classification has a long history in machine
learning~\cite{elkan2001foundations}, but most work focuses on
misclassification costs in fixed, single-model settings. We generalize this setting to sequential pipelines over multiple models, treating both computational and human costs explicitly.
 
\paragraph{Hierarchical reinforcement learning.}
Hierarchical RL~\cite{barto2003recent} decomposes a long-horizon problem
into a hierarchy of sub-goals, enabling structured exploration.  Option
frameworks~\cite{sutton1999between} provide a principled abstraction for
this decomposition.  We adapt this to document classification, where
options correspond to classes of evidence-gathering actions.
 
\paragraph{Adaptive inference.}
Early-exit networks~\cite{teerapittayanon2016branchynet} and cascaded
classifiers reduce inference cost by routing easy examples to shallow
sub-networks.  \emph{DocHRL} generalises this idea to a heterogeneous pool of
classifiers, OCR engines, LLMs, and human reviewers, and learns the
routing policy via RL rather than hardcoding thresholds.

\paragraph{PPO variants.}
Proximal Policy Optimization (PPO)~\cite{schulman2017proximal} enforces
a trust region by clipping importance ratios. Several works replace this
hard clip with smooth alternatives: PPOS~\cite{zhu2020proximal} applies
a $\tanh$ penalty outside the clip region, PPO-ALR~\cite{su2024policy}
bounds the ratio via $\tanh(r-1)+1$, SAPO~\cite{li2025soft} uses a
sigmoid gate with asymmetric temperatures, and
PSPO~\cite{dwyer2025not} linearly interpolates old and current
probabilities for non-vanishing gradients. Our variant uses a
$\tanh$-gated ratio with a linear leak term, providing a smooth trust
region while guaranteeing non-zero gradient for extreme ratios.

\paragraph{Adaptive return normalization.}
Normalizing returns stabilises value learning.  Pop-Art~\cite{vanhasselt2016learning} tracks running mean and variance of returns; DreamerV3~\cite{hafner2023mastering} introduced $\operatorname{symlog}(x)=\operatorname{sign}(x)\log(1+|x|)$ as a fixed squashing function paired with two-hot value encoding. Rational Reward Shaping~\cite{mcinnis2025adapting} learns a ratio of polynomials to transform rewards, while percentile normalization~\cite{hafner2023mastering} scales by the P5--P95 range. Our AdaPolyTan blends a polynomial and a $\tan$ transform with parameters that adapt to the mean return magnitude, producing a stateless, functional normalisation matched to a bounded value head.

\section{Methodology}
\label{sec:meth}
 
\subsection{Cost Models for Document Classification Pipelines}
 
We study four pipeline architectures and define their total expected cost
$C_t$ per classification instance in terms of three components:
 
\begin{itemize}
  \item $C_i$: \textit{inference cost} - the computational (monetary)
        cost of a single model forward-pass, fixed per model and hardware.
  \item $C_f$: \textit{cost of failure} - the expected downstream
        business loss when a document is assigned the wrong label.
        Assumed fixed across classes; treated as a hyperparameter.
  \item $C_h$: \textit{human labeller cost} - the fully-loaded cost of
        manual expert review.  Fixed per labeller.
\end{itemize}
 
\noindent\textbf{Standalone classifier:}
\begin{equation}
  C_t = C_i + E_r \cdot C_f,
  \label{eq:standalone}
\end{equation}
where $E_r = 1 - A$ is the error rate and $A$ is classification accuracy.
 
\noindent\textbf{Ensemble pipeline:}
\begin{equation}
  C_t = \sum_k C_i^{(k)} + \bar{E}_r \cdot C_f,
  \label{eq:ensemble}
\end{equation}
where $\bar{E}_r$ is the average error rate of the majority-voted ensemble.
 
\noindent\textbf{Human-in-the-loop pipeline:}
\begin{equation}
  C_t = C_i + (1 - P_h)\cdot E_r \cdot C_f + P_h \cdot C_h + P_h \cdot E_h \cdot C_f,
  \label{eq:hitl}
\end{equation}
where $P_h$ is the probability of escalation to a human and $E_h$ is
the human error rate.  The last term accounts for the expected cost of
residual human error.

\begin{figure}[htbp]
  \centering
  \includegraphics[width=0.4\textwidth]{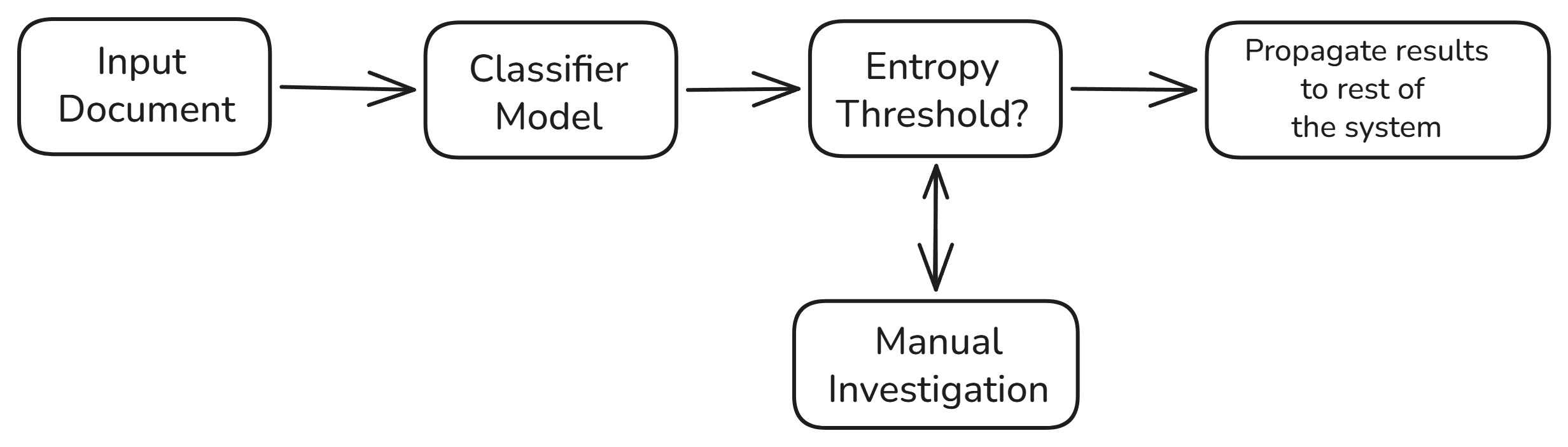}
  \caption{Human-in-the-loop pipeline with entropy-based escalation. The classifier processes an input; if the normalized Shannon entropy $\mathcal{H}$ exceeds threshold $T_h$, the sample is escalated to a human operator. Otherwise, the classifier makes the final decision.}
  \label{fig:hitl_pipe}
\end{figure}

\noindent\textbf{Entropy-based escalation.}
Using the normalised Shannon entropy of the classifier's output distribution,
\begin{equation}
  \mathcal{H} = -\sum_{i=0}^{K-1} P(y_i)\,\frac{\log P(y_i)}{\log K},
  \label{eq:entropy}
\end{equation}
the escalation probability $P_h \equiv P(\mathcal{H} > T_h)$ and the classifier error rate $E_r \equiv 1 - A(T_h)$ become functions of the threshold $T_h$, as illustrated in Figure~\ref{fig:hitl_pipe}. Eq.~\eqref{eq:hitl} can then be minimized by numerically solving $\partial C_t / \partial T_h = 0$.
 
\noindent\textbf{\emph{DocHRL} pipeline.}
Unlike the above, \emph{DocHRL} makes $C_t$ adaptive per instance by learning a
policy that selects the cheapest \emph{sufficient} set of actions for each
document.  The \emph{DocHRL} objective is to minimise $\mathbb{E}[C_t]$ over the
document distribution.
 
\subsection{Cost Quantification}
\label{sec:cost_quant}
 
\textbf{Inference cost $C_i$.}
We compute $C_i = R_\text{GPU} / T_\text{inf}$, where $R_\text{GPU}$ is
the on-demand GPU hourly rate and $T_\text{inf}$ is model throughput
(images/hr).  All models operate on $384 \times 384$ pixel inputs.
Since direct benchmarks at this resolution are scarce, throughput figures
at $1024 \times 1024$ are scaled by a factor of $7.0$ (ratio of pixel
counts).  GPU pricing used: T4 at \$0.15/hr, A100 at \$0.66/hr,
L4 at \$0.70/hr, and A4000 at \$0.25/hr.
Table~\ref{tab:ci} summarises the estimated $C_i$ values.

\begin{table}[h!]
\centering
\small
\setlength{\tabcolsep}{3.5pt}
\begin{tabular}{llr}
\toprule
\rowcolor{blue!10}
\textbf{Policy / Action} & \textbf{Hardware} & \boldmath$\Ci$ \textbf{(\$/doc)} \\
\midrule
EfficientNet-V2-S     & T4 GPU     & $5.14 \times 10^{-9}$ \\
EfficientNet-V2-M     & T4 GPU     & $6.56 \times 10^{-9}$ \\
ResNet-18             & A4000 GPU  & $8 \times 10^{-8}$ \\
ResNet-50             & T4 GPU     & $9.26 \times 10^{-8}$ \\
SWIN-small            & A4000 GPU  & $7 \times 10^{-8}$ \\
SWIN-B                & T4 GPU     & $1.7 \times 10^{-7}$ \\
Donut                 & A4000 GPU  & $1.4 \times 10^{-6}$ \\
Tesseract (CPU OCR)   & CPU        & $1.4 \times 10^{-6}$ \\
doctr (GPU OCR)       & T4 GPU     & $1.5 \times 10^{-4}$ \\
Mistral OCR       & API     & $1 \times 10^{-3}$ \\
Gemini-2.5 Flash      & API        & $1.7 \times 10^{-3}$ \\
GPT-4o-mini         & API        &  $4 \times 10^{-3}$  \\
Human labeller        & Human      & $2.25$ \\
\bottomrule
\end{tabular}
\vspace{7pt}
\caption{Estimated inference cost $\Ci$ per document (selected entries). 
Foundational model costs are averaged over the RVL-CDIP dataset for image-based classification. 
For foundational and OCR models (Tesseract, doctr, Gemini, GPT-4o-mini), input images are $960 \times 640$ pixels. 
All other classifier benchmarks at $384\!\times\!384$ pixel resolution.}
\label{tab:ci}
\end{table}
 
A striking observation is the near cost-parity between a highly-optimised
deep learning classifier (EfficientNet-V2-S at 
$\approx\!5\!\times\!10^{-9}$\,\$/doc)
and a CPU-bound OCR engine (Tesseract at $\approx\!1.4\!\times\!10^{-6}$\,\$/doc).
Human labelling is four to seven orders of magnitude more expensive than any
automated policy, which creates a dominant economic incentive for \emph{DocHRL}
to minimise escalation.
 
\textbf{Cost of failure $C_f$.}
Following standard practice in cost-sensitive
learning~\cite{elkan2001foundations}, we fix $C_f = 100.0$ (normalised
units) across all document classes and experiments.  This value represents
the expected downstream business loss from a misclassification and is
treated as a tunable hyperparameter.
 
\textbf{Human labeller cost $C_h$.}
Assuming a fully-loaded rate of \$27/hr and an average review time of
five minutes per document:
\begin{equation}
  C_h = \frac{27.00}{3600} \times 300 = \$2.25 \text{ per document.}
\end{equation}

Together, $C_i$, $C_f$, and $C_h$ fully parameterise the cost models of Eqs.~\eqref{eq:standalone}--\eqref{eq:hitl}; their instantiated values are used to compute $C_t$ for all baseline systems in Section~\ref{sec:results} (Table~\ref{tab:cost_compare}).

\textbf{Human-in-loop system cost}

\begin{table}[h!]
\centering
\setlength{\tabcolsep}{3.5pt}
\begin{tabular}{lrrrr}
\toprule
\rowcolor{blue!10}
\textbf{Model} & \boldmath$T_{\mathrm{min}}$ & \boldmath$C_t^{\mathrm{min}}$ & \boldmath$T_{\mathrm{max}}$ & \boldmath$C_t^{\mathrm{max}}$ \\
\midrule
ResNet-18     & 0.105  & 3.345 & 0.999 & 12.502 \\
ResNet-50     & 0.0739 & 3.237 & 0.931 & 10.639 \\
SWIN-small    & 0.0378 & 3.242 & 0.755 & 9.658  \\
SWIN-B       & 0.0501 & 3.167 & 0.845 & 9.772  \\
EfficientNet-V2-S & 0.0385 & 3.102 & 0.776 & 8.908  \\
EfficientNet-V2-M  & 0.0317 & 3.085 & 0.767 & 8.668  \\
\bottomrule
\end{tabular}
\vspace{7pt}
\caption{Minimum-cost and maximum-cost entropy thresholds for each HIL model, obtained by numerically solving $\partial C_t / \partial T_h = 0$ (Eq.~\ref{eq:hitl}) using the test-set entropy distribution. $T_{\mathrm{min}}$ is the threshold that minimises total cost $C_t$; $T_{\mathrm{max}}$ is the threshold that maximises it. Both are derived per model under $C_f = 100$, $C_h = 2.25$, $E_h = 0.04$.}
\label{tab:cost_minmax}
\end{table}

Table~\ref{tab:cost_minmax} includes only models that output a full probability distribution over classes, as entropy computation (Eq.~\ref{eq:entropy}) requires this. LLMs that output discrete labels and OCR models are therefore excluded from this analysis; OCR models (e.g.\ doctr, Tesseract) appear in Table~\ref{tab:ci} solely as \emph{components} invoked by \emph{DocHRL}, not as standalone classifiers. For each model, two operating points are derived by numerically solving $\partial C_t / \partial T_h = 0$ (Eq.~\eqref{eq:hitl}) over the test-set entropy distribution with $E_h = 0.04$. This yields $T_{\mathrm{min}}$, the threshold that \emph{minimises} total cost, and $T_{\mathrm{max}}$, the threshold that \emph{maximises} it, along with their corresponding costs $C_t^{\mathrm{min}}$ and $C_t^{\mathrm{max}}$, as reported in Table~\ref{tab:cost_minmax}. The existence of $T_{\mathrm{max}} < 1$ is noteworthy: it arises because only a small fraction of samples are classified with very low confidence (high entropy), so near $T_{\mathrm{max}}$ the system almost mirrors standalone classifier behaviour while routing a handful of documents to the human labeller, incurring marginally higher total cost. Human-in-the-loop pipelines at $T_{\mathrm{min}}$ consistently outperform both standalone models and ensembles.

\subsection{The \emph{DocHRL} Framework}
\label{sec:dochrl}

\begin{figure*}[t]
  \centering
  \includegraphics[width=\textwidth]{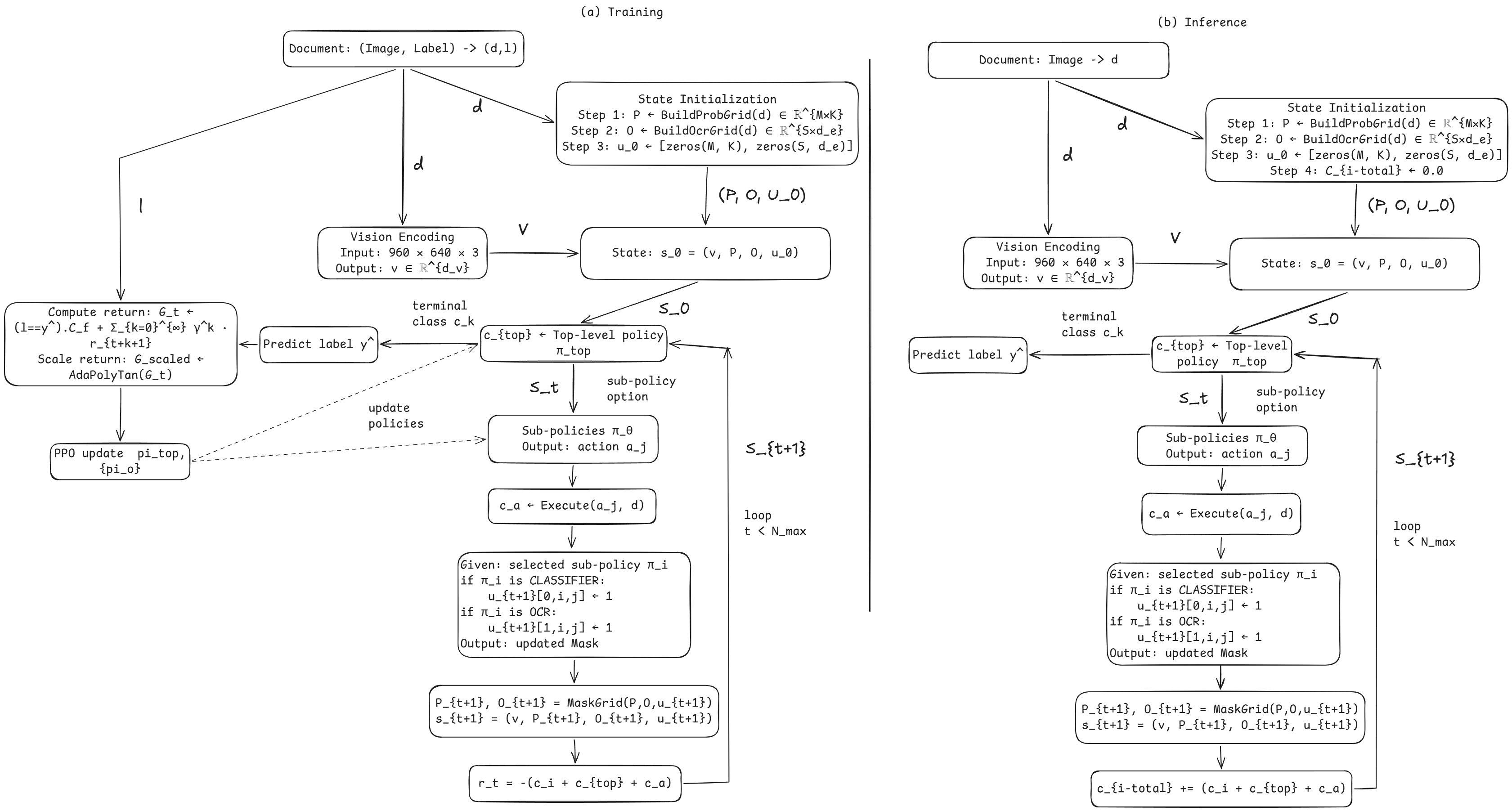}
  \caption{Overview of the \emph{DocHRL} framework. \textbf{(a) Training:}
    the agent interacts with each document for at most $N_\mathrm{max}$
    steps, accumulates cost $C_t$, and updates both policies via a custom
    PPO variant with tanh-gated trust region and AdaPolyTan reward scaling.
    \textbf{(b) Inference:} the same
    hierarchy runs greedily ($\operatorname{argmax}$) with no reward computation.}
  \label{fig:arch}
\end{figure*}

Figure~\ref{fig:arch} provides an overview of the \emph{DocHRL} architecture,
with training and inference stages shown in separate panels.

\textbf{State representation.}
The agent state at step $t$ is $s_t = (v,\, P_t,\, O_t,\, u_t)$, where
$v \in \mathbb{R}^{d_v}$ is a vision embedding of the document (extracted
once by a frozen \texttt{VisionEncoder} at episode start);
$P_t \in [0,1]^{M \times K}$ is a probability grid accumulating soft-label
predictions from all policy actions invoked so far, with $M$ classification-
oriented actions (classifier variants, LLMs, human annotators) and $K=16$
classes; $O_t \in \mathbb{R}^{S \times d_e}$ is an OCR grid containing text
embeddings from $S$ OCR sources; and $u_t \in \{0,1\}^M$ is a binary mask
tracking which actions have already been used, enabling the agent to avoid
redundant queries. At episode start, $P_0$ and $O_0$ are initialised as
zero grids via \texttt{BuildProbGrid} and \texttt{BuildOcrGrid}
respectively, and $u_0 = \mathbf{0}^M$; grids are updated in-place as
sub-policies execute actions. Before each policy query,
\texttt{MaskGrids}$(P_0, O_0, u_t)$ zeros out rows corresponding to
already-used actions, preventing the policies from re-querying them.

\textbf{Policy hierarchy.}
\emph{DocHRL} employs two policy levels.  The \emph{top-level policy}
$\pi_\mathrm{top}$ selects an option from
$\mathcal{O} = \{\texttt{classify}, \texttt{llm}, \texttt{ocr},
\texttt{human}, c_1,\ldots,c_K\}$, where the first four options delegate
to a sub-policy and $c_1,\ldots,c_K$ are terminal actions that
immediately predict a class label.  Each non-terminal option $\theta \in
\mathcal{O}$ has a dedicated \emph{sub-level policy} $\pi_\theta(s_t)$ with
its own action set $\mathcal{A}_\theta$: for \texttt{classify}, these are
specific classifier models; for \texttt{ocr}, specific OCR engines;
for \texttt{llm}, specific language models; and for \texttt{human},
annotators of different accuracy levels.

All policies share a common architecture that fuses the three observation modalities, prob grid, OCR grid, and vision embedding, through dedicated encoders followed by a shared MLP, and outputs both action logits and a scalar value estimate. The value head is a two-layer MLP whose output passes through $0.5 \cdot \tanh$, bounding $V_\theta(s)$ to $[-0.5, 0.5]$ to match the AdaPolyTan-normalised return range.  Top and sub-level agents differ only
in their encoder widths and action-space sizes; sub-level agents use
smaller internal dimensions since their decision scope is narrower.

\textbf{Reward function.}
At each step $t$, the agent incurs a cost depending on the action taken.
Let $\mathcal{T}$ be the set of steps where the top agent acts, and
$\mathcal{S}$ the set of steps where a sub-policy executes an action.
For a top-agent step $t \in \mathcal{T}$, the cost is its inference cost:
\begin{equation}
  \kappa_t = C_{\pi}^{\mathrm{top}},
\end{equation}
where $C_{\pi}^{\mathrm{top}}$ is a nominal fee for querying the top
policy.  If the top agent selects a terminal class $\hat{y}$ at step $t$,
a failure penalty is added:
\begin{equation}
  \kappa_t = C_{\pi}^{\mathrm{top}} + C_f \cdot \mathbf{1}[\hat{y} \neq y],
\end{equation}
where $C_f$ is the cost of an incorrect classification.

If the top agent instead delegates to a sub-policy $o \in
\{\texttt{classify}, \texttt{llm}, \texttt{ocr}, \texttt{human}\}$,
the corresponding sub-policy $\pi_o$ selects an action $a \in
\mathcal{A}_o$ at the subsequent step $t'$, incurring:
\begin{equation}
  \kappa_{t'} = C_{\pi}^{\mathrm{sub}} + C_{\mathrm{exec}}(a),
\end{equation}
where $C_{\pi}^{\mathrm{sub}}$ is the sub-policy inference cost and
$C_{\mathrm{exec}}(a)$ is the execution cost of action $a$ (e.g., a
classifier inference, an LLM API call, or a human annotator fee).

The total episode cost accumulates over all $T$ steps:
\begin{equation}
  C_{\mathrm{total}} = \sum_{t=1}^{T} \kappa_t,
\end{equation}
and the scalar reward is its negation, assigned at episode termination and
propagated to all transitions via discounted returns:
\begin{equation}
  R_t = -\sum_{k=t}^{T} \gamma^{k-t} \kappa_k.
\end{equation}

\textbf{PPO variant.}
We replace standard PPO's hard importance-ratio clipping with a smooth
$\tanh$-gated trust region.  Given the importance ratio
$r_t = \exp(\log\pi_\theta(a_t|s_t) - \log\pi_{\theta_\mathrm{old}}(a_t|s_t))$,
the policy loss is:
\begin{equation}
  L^\mathrm{policy} = -\mathbb{E}\!\left[
    \bigl(\beta \tanh(r_t-1) + 1 + \alpha(r_t-1)\bigr) \cdot A_t
  \right],
  \label{eq:tanh_gate}
\end{equation}
where $\beta = 0.2$ controls the asymptotic bound $[1-\beta, 1+\beta]$,
and $\alpha = 0.01$ is a linear leak term that preserves non-zero gradient
for extreme ratios---avoiding the hard gradient cutoff of standard
clipping.  The total PPO loss is:
\begin{equation}
  L^\mathrm{total} = L^\mathrm{policy} + c_1 L^\mathrm{value} - c_2(t) H(\pi),
  \label{eq:total_loss}
\end{equation}
with value loss coefficient $c_1 = 0.5$ and an entropy coefficient
$c_2(t)$ that decays linearly from $0.2$ to $0.02$ over training to
anneal exploration.

\textbf{AdaPolyTan return scaling.}
Raw returns span a wide dynamic range, which can destabilise PPO training.
We apply a custom \emph{AdaPolyTan} transformation before computing value
loss and advantages.  Given a batch of raw returns $x$ with expected range
$[v_\mathrm{min}, v_\mathrm{max}]$:
\begin{align}
  \tilde{x} &= \frac{\operatorname{clamp}\!\bigl(\operatorname{mean}(x)\bigr) - v_\mathrm{min}}{v_\mathrm{max} - v_\mathrm{min}}, \\
  p         &= \bigl\lfloor p_\mathrm{max} - (p_\mathrm{max}-p_\mathrm{min})\,\tilde{x}\bigr\rfloor, \\
  \beta_b   &= \beta_\mathrm{min} + (\beta_\mathrm{max}-\beta_\mathrm{min})\,\tilde{x}, \\
  x_s       &= 2\,\frac{x - v_\mathrm{min}}{v_\mathrm{max} - v_\mathrm{min}} - 1 \;\in\; [-1, 1], \\
  y         &= \operatorname{clamp}\!\left(
                  \tfrac{1}{2}\bigl[\beta_b\, x_s^{2p+1}
                  + (1-\beta_b)\,\widetilde{\tan}(x_s)\bigr],
                  -0.5,\, 0.5
                \right),
\end{align}
where $\widetilde{\tan}$ is $\tan$ linearly rescaled to $[-1, 1]$.
The parameters adapt per batch based on the normalised mean return
$\tilde{x}$: $p \in [100, 1000]$ and $\beta_b \in [0.2, 0.5]$.  When
returns are poor (low $\tilde{x}$), a high $p \approx 1000$ produces a
near-binary polynomial that sharply separates costly from cheap episodes,
while a low $\beta_b \approx 0.2$ keeps the output smooth via the
$\tan$ component.  As training improves (high $\tilde{x}$), $p$ drops
to $100$, softening the polynomial, and $\beta_b$ rises to $0.5$, giving
equal weight to both components for finer gradient resolution near
convergence.

\textbf{Training.}
Each episode processes a single labelled document.  The agent invokes at
most $N_\mathrm{max}=14$ steps; at each step, either the top agent selects
a terminal class (ending the episode) or delegates to a sub-policy, which
selects a specific action and updates the state.  At episode end, returns
are computed via $\gamma$-discounted cumulative cost, normalised through
AdaPolyTan, and both $\pi_\mathrm{top}$ and all sub-policies are updated
independently via the PPO variant described above.  If the agent exhausts
its step budget without selecting a terminal label, $\pi_\mathrm{top}$ is
queried one final time with all non-terminal options masked, forcing a
terminal prediction.

\textbf{Inference.}
At test time, the same loop runs with greedy $\operatorname{argmax}$
selection rather than Boltzmann sampling, and no reward computation or
policy update is performed.
\section{Experiments}
\label{sec:exp}
 
\subsection{Dataset}
 
We train and evaluate on the \textbf{RVL-CDIP} dataset~\cite{harley2015evaluation},
a benchmark consisting of 400,000 grayscale document images spanning 16
categories (letter, form, email, handwritten, advertisement, scientific
report, scientific publication, specification, file folder, news article,
budget, invoice, presentation, questionnaire, resume, and memo), with
25,000 images per class. We employ a standard 80:10:10 split for the training, validation, and test sets throughout our experiments. All images are
resized to $960 \times 640$ pixels prior to processing.

\subsection{Evaluation Protocol}
\label{sec:eval_protocol}
 
We evaluate \emph{DocHRL} along two complementary axes.
 
\textbf{Classification performance.}
We report per-class accuracy and macro-averaged precision, recall, and F1
on the standard RVL-CDIP test split (33,669 images).
\emph{DocHRL} is compared against seven standalone classifiers including
ResNet-50~\cite{he2016deep},
EfficientNet-V2-S and V2-M~\cite{tan2021efficientnetv2},
SWIN-small and SWIN-B~\cite{liu2021swin},
Gemini-2.5 Flash, and GPT-4o-mini as well as two ensemble baselines: a majority-voting ensemble of all
CNN/ViT classifiers (\textit{CNN Ensemble}) and a soft-voting
ensemble of the two LLMs (\textit{LLM Ensemble}).
 
\textbf{Cost efficiency.}
For each standalone baseline the average per-document total cost is
estimated via Eq.~\eqref{eq:standalone}, with $C_f = 100$ (normalised
units) and $C_i$ omitted as it is at least four orders of magnitude
smaller than the failure cost term (see Table~\ref{tab:ci}).
For \emph{DocHRL}, $\bar{C}_t = 2.74$ is measured directly by averaging the
accumulated episode cost $C_t$ over all test documents.

\subsection{Results}
\label{sec:results}

\subsubsection{Cost Efficiency}
 
Table~\ref{tab:cost_compare} compares the average per-document total
cost $\bar{C}_t$ across all systems.  \emph{DocHRL} achieves an average cost
of \textbf{2.74 normalised units}, representing reductions of
\textbf{69\%} relative to the best standalone classifier
(EfficientNet-V2-M: 8.74), \textbf{65\%} relative to the CNN ensemble
(7.74), \textbf{88\%} relative to the best LLM standalone
(Gemini-2.5 Flash: 23.29).

\begin{table}[h!]
\centering
\setlength{\tabcolsep}{3.5pt}
\begin{tabular}{lcc}
\toprule
\rowcolor{blue!10}
\textbf{System} & \textbf{Macro Acc.} $\uparrow$ & \boldmath$\bar{C}_t$ $\downarrow$ \\
\midrule
GPT-4o-mini (standalone)       & 0.714 & 28.64 \\
Gemini-2.5 Flash (standalone)  & 0.767 & 23.29 \\
LLM Ensemble                   & 0.843 & 15.66 \\
ResNet-50 (standalone)         & 0.893 & 10.72 \\
SWIN-small (standalone)        & 0.902 &  9.79 \\
SWIN-B (standalone)            & 0.903 &  9.67 \\
EfficientNet-V2-S (standalone) & 0.910 &  8.99 \\
EfficientNet-V2-M (standalone) & 0.913 &  8.74 \\
CNN Ensemble                   & 0.923 &  7.74 \\
\midrule
\textbf{\emph{DocHRL} (ours)}  & \textbf{0.973} & \textbf{2.74} \\
\bottomrule
\end{tabular}
\vspace{7pt}
\caption{Average per-document total cost $\bar{C}_t$ (normalised units,
  $C_f\!=\!100$) and macro accuracy.
  Standalone costs are computed via Eq.~\eqref{eq:standalone} and
  ensemble costs via Eq.~\eqref{eq:ensemble}, with $C_i$ omitted as
  negligible. \emph{DocHRL} cost is measured directly from test-set
  episodes. $\downarrow$ lower is better for cost; $\uparrow$ higher
  is better for accuracy. CNN Ens.\ = majority-voting over
  ResNet-50, EfficientNet-V2-S, EfficientNet-V2-M, SWIN-small, and
  SWIN-B. LLM Ens.\ =
  soft-voting over Gemini-2.5 Flash, and
  GPT-4o-mini.}
\label{tab:cost_compare}
\end{table}
 
The cost reduction achieved by \emph{DocHRL} is not simply a consequence of
routing all documents to a cheaper model uniformly, which would
require trading away accuracy.  Instead, \emph{DocHRL} simultaneously achieves
the \emph{lowest} cost \emph{and} the \emph{highest} accuracy of any
system evaluated.  This is possible because the agent routes easy,
high-confidence documents directly to inexpensive lightweight
classifiers, while reserving OCR, LLM, and human-review actions for
documents where initial classifier confidence is low.
The document classes that benefit most from this adaptive routing -
\textit{form}, \textit{scientific report}, and \textit{presentation} -
are precisely those where all standalone classifiers show the largest
accuracy deficits, confirming that \emph{DocHRL} is learning a meaningful
difficulty-aware routing policy rather than collapsing to a single
preferred action. Compared separately to the best human-in-the-loop configuration, \emph{DocHRL} also outperforms the optimal HIL system: EfficientNet-V2-M at its cost-minimising threshold $T_{\mathrm{min}} = 0.0317$ achieves $C_t^{\mathrm{min}} = 3.085$ (Table~\ref{tab:cost_minmax}), whereas \emph{DocHRL} achieves $\bar{C}_t = 2.74$, a further \textbf{11.1\%} reduction, while also delivering superior accuracy (0.973 vs.\ 0.913 standalone).

\begin{table*}[t]
\centering
\setlength{\tabcolsep}{3.5pt}
\begin{tabular}{lcccccccccc}
\toprule
\rowcolor{blue!10}
\textbf{Class} & \textbf{RN-50} & \textbf{Eff-S} & \textbf{Eff-M} & \textbf{SW-s} & \textbf{SW-B} & \textbf{Gem.} & \textbf{GPT} & \textbf{CNN Ens.} & \textbf{LLM Ens.} & \textbf{\emph{DocHRL}} \\
\midrule
Letter               & 0.874 & 0.909 & 0.910 & 0.876 & 0.910 & 0.696 & 0.733 & 0.924 & 0.707 & \textbf{0.954} \\
Form                 & 0.854 & 0.799 & 0.837 & 0.830 & 0.827 & 0.646 & 0.500 & 0.848 & 0.815 & \textbf{0.928} \\
Email                & 0.984 & 0.986 & 0.990 & 0.990 & 0.989 & 0.895 & 0.872 & 0.991 & 0.891 & \textbf{1.000} \\
Handwritten          & 0.959 & 0.958 & 0.946 & 0.960 & 0.958 & 0.781 & 0.263 & 0.962 & 0.500 & \textbf{0.975} \\
Advertisement        & 0.895 & 0.925 & 0.934 & 0.903 & 0.933 & 0.881 & 0.884 & 0.936 & 0.974 & \textbf{0.978} \\
Scientific report    & 0.754 & 0.820 & 0.795 & 0.813 & 0.731 & 0.749 & 0.702 & 0.830 & 0.765 & \textbf{0.935} \\
Scientific pub.\     & 0.918 & 0.925 & 0.932 & 0.921 & 0.909 & 0.909 & \textbf{0.969} & 0.930 & 0.966 & \textbf{0.985} \\
Specification        & 0.903 & 0.937 & 0.939 & 0.933 & 0.922 & 0.780 & 0.658 & 0.941 & 0.833 & \textbf{0.977} \\
File folder          & 0.947 & 0.946 & 0.949 & 0.955 & 0.955 & 0.841 & 0.628 & 0.968 & 0.964 & \textbf{0.981} \\
News article         & 0.912 & 0.892 & 0.916 & 0.909 & 0.911 & 0.898 & 0.978 & 0.916 & \textbf{1.000} & 0.985 \\
Budget               & 0.871 & 0.907 & 0.899 & 0.886 & 0.907 & 0.587 & 0.455 & 0.920 & 0.586 & \textbf{0.970} \\
Invoice              & 0.900 & 0.924 & 0.923 & 0.899 & 0.910 & 0.787 & 0.884 & 0.923 & 0.861 & \textbf{0.969} \\
Presentation         & 0.835 & 0.872 & 0.877 & 0.840 & 0.844 & 0.335 & 0.243 & 0.886 & 0.444 & \textbf{0.962} \\
Questionnaire        & 0.808 & 0.891 & 0.880 & 0.859 & 0.848 & 0.806 & 0.791 & 0.889 & 0.825 & \textbf{1.000} \\
Resume               & 0.960 & 0.947 & 0.940 & 0.941 & 0.969 & 0.801 & 0.969 & 0.962 & \textbf{1.000} & \textbf{1.000} \\
Memo                 & 0.907 & 0.923 & 0.933 & 0.914 & 0.929 & 0.880 & 0.897 & 0.935 & \textbf{1.000} & 0.965 \\
\midrule
\textbf{Macro acc.}  & 0.893 & 0.910 & 0.913 & 0.902 & 0.903 & 0.767 & 0.714 & 0.923 & 0.843 & \textbf{0.973} \\
\textbf{Macro F1}    & 0.893 & 0.910 & 0.913 & 0.902 & 0.903 & 0.767 & 0.703 & 0.923 & 0.826 & \textbf{0.973} \\
\bottomrule
\end{tabular}
\vspace{7pt}
\caption{Per-class accuracy and macro-averaged metrics on the RVL-CDIP
  test set.  Best result per class in \textbf{bold}. CNN Ens.\ = majority-voting over ResNet-50, EfficientNet-V2-S, EfficientNet-V2-M, SWIN-small, and SWIN-B. LLM Ens.\ =
  soft-voting over Gemini-2.5 Flash, and
  GPT-4o-mini.}
\label{tab:cls_compare}
\end{table*}

\subsubsection{Classification Performance}
 
Table~\ref{tab:cls_compare} reports macro-averaged and per-class accuracy
for all systems.  \emph{DocHRL} achieves the highest macro accuracy of
\textbf{97.27\%} and macro F1 of \textbf{0.973}, outperforming the best
standalone classifier (EfficientNet-V2-M at 91.26\%) by \textbf{6.0
percentage points} and the strongest ensemble (CNN Ensemble at 92.26\%) by
\textbf{5.0 percentage points}.
 
\emph{DocHRL} reaches perfect per-class accuracy on \textit{email},
\textit{questionnaire}, and \textit{resume}.
The most challenging classes across all systems are \textit{form} and
\textit{scientific report}, which are visually similar to several
neighbouring categories.  Even on these, \emph{DocHRL} substantially
outperforms all baselines, suggesting that the adaptive combination of
visual, OCR, and LLM evidence is particularly beneficial for ambiguous
document types.
 
LLM-based classifiers (Gemini-2.5 Flash: 76.7\%, GPT-4o-mini: 71.4\%)
perform well below all vision-based methods, consistent with prior
findings that zero-shot image-level document classification is not a
strong suit of general-purpose LLMs without fine-tuning.

\textbf{Per-class cost analysis.}
The cost reduction is not uniform across document classes.
Visually unambiguous classes such as \textit{email}, \textit{questionnaire},
and \textit{resume} consistently incur low average cost, reflecting the
agent's tendency to terminate quickly after a single cheap classifier
invocation - all three reach perfect accuracy with minimal effort.
By contrast, \textit{form}, \textit{scientific report}, and
\textit{presentation} incur higher average cost per episode, indicating
that the agent invokes additional OCR or LLM evidence before committing
to a label on these ambiguous classes.
Crucially, this is also where \emph{DocHRL} shows the largest accuracy gains over
standalone baselines: \textit{form} improves from 0.854 (best standalone)
to 0.928, \textit{scientific report} from 0.820 to 0.935, and
\textit{presentation} from 0.877 to 0.962.
This alignment between high episode cost and high accuracy gain confirms
that \emph{DocHRL} is learning a meaningful difficulty-aware routing policy,
spending more only where it matters.

\section{Conclusion}
\label{sec:conc}

We presented \emph{DocHRL}, a hierarchical reinforcement learning framework that
reframes document classification as a cost-aware sequential decision
problem.  By incorporating inference cost, misclassification cost, and
human-in-the-loop cost directly into the reward signal, and training a
two-level policy hierarchy with PPO and adaptive reward scaling, \emph{DocHRL}
learns to allocate computational resources in proportion to the difficulty
of each document instance.

Evaluated on the RVL-CDIP benchmark against six standalone classifiers
and two ensemble baselines, \emph{DocHRL} achieves the best classification
performance of any system (macro F1 = 0.973, +6.0 pp over the best
standalone) while simultaneously reducing the average per-document cost
by 69\% relative to the best standalone classifier and 65\% relative to
the best ensemble.  These results demonstrate that cost-aware reinforcement
learning is a practical and principled approach to building economically
efficient document understanding systems, and that accuracy and cost
efficiency need not be traded against one another when the routing policy
is learned rather than fixed.

\paragraph{Limitations and future work.}
We emphasise that \emph{DocHRL} is a \textit{proof-of-concept} for
cost-aware adaptive routing in document classification pipelines,
targeting scenarios where the ensemble of heterogeneous classifiers including LLMs and human-in-the-loop review is already
operationally justified, and where minimising total expected cost
$C_t$ is as important as, or more important than, raw classification
accuracy alone.  Within this setting, the results establish that
hierarchical RL can learn a routing policy that simultaneously
improves both objectives.

Several directions remain for future work. First, the current work assumes a fixed, domain-independent failure
cost $C_f$.  In practice, the relative importance of cost and accuracy
varies by deployment context; learning or estimating class-specific
$C_f$ values from downstream feedback would allow the agent to be
more aggressive on low-stakes classes and more cautious on high-stakes
ones, making the framework directly applicable to regulated domains
such as healthcare or legal document processing.

Second, the human labeler in our experiments is implemented via a
stochastic function that emits correct labels with probability
$1 - E_h$ and incorrect ones otherwise, rather than modelling
realistic human behaviour such as fatigue, learning effects, or
response time variability.  While this suffices for a proof-of-concept
cost--accuracy trade-off analysis, future work should incorporate
behavioural models of human annotators or, ideally, deploy the system
with actual human-in-the-loop feedback to validate the cost savings
under real-world conditions.

Third, a natural question is whether a simpler learned cascade such as a confidence-threshold router from a cheap classifier to a
stronger one could achieve comparable cost reductions with less
training complexity.  We view such cascades as a useful ablation
baseline for future work, and expect that the full \emph{DocHRL} framework
will show the largest gains precisely in the complex, multi-policy
settings it is designed for, where a two-stage cascade cannot capture
the full range of available actions.

Fourth, regarding scalability: the hierarchical PPO training loop
requires up to $N_{\max}$ model invocations per training document,
and the action space grows with the classifier pool.  For the
proof-of-concept scale evaluated here this is tractable, but
scaling to larger pools or higher-volume pipelines will benefit from
off-policy RL or experience replay to reduce training cost.
Batch-level scheduling where the agent jointly routes a stream
of documents rather than treating each independently is a
further avenue for improving throughput efficiency.

Finally, the current evaluation uses the balanced RVL-CDIP benchmark.
Understanding \emph{DocHRL}'s behaviour under real-world class imbalance,
distribution shift, and varying $C_f$ regimes is an important next
step toward production deployment.

\newpage

\bibliography{reference_file}
\bibliographystyle{icml2026}




\end{document}